\documentclass{article}

\usepackage{arxiv}

\usepackage{libertine}
\usepackage{graphicx}
\usepackage{array,booktabs}
\usepackage{subfig}
\usepackage{pdflscape}
\usepackage{pdfpages}
\usepackage{hyperref}
\newcolumntype{P}[1]{>{\centering\arraybackslash}p{#1}}

\title{Trust in Language Grounding: a new AI challenge for human-robot teams}
\author{David M. Bossens\\ University of Southampton\\ d.m.bossens@soton.ac.uk \And Christine Evers\\ University of Southampton\\ c.evers@soton.ac.uk}

\begin{document}

\maketitle

\begin{abstract}
The challenge of language grounding is to fully understand natural language by grounding language in real-world referents. While AI techniques are available, the widespread adoption and effectiveness of such technologies for human-robot teams relies critically on user trust.  This survey provides three contributions relating to the newly emerging field of \textit{trust in language grounding}, including a) an overview of language grounding research in terms of AI technologies, data sets, and user interfaces; b) six hypothesised trust factors relevant to language grounding, which are tested empirically on a human-robot cleaning team; and c) future research directions for trust in language grounding.
\end{abstract}


\maketitle

\section{Introduction}
Human-robot teams provide a promising prospect for society by harnessing the complementary strengths of humans and robots. The robot can automate mundane, dangerous, or difficult tasks that humans do not wish to do while the human can devise high-level plans and instruct, guide, or correct the robot when needed. In a variety of applications, there is a need for continual communication between human and robot to inform task requirements changing over time, point out contextual nuances, such as unexpected obstacles or circumstances, and to engage the human. In these cases, communication via natural language, often in conjunction with non-verbal cues, is an attractive option for a variety of applications such as care \cite{Vincze2014,Hoffman2015},  assisted surgery \cite{Jacob2012}, and space teleoperation \cite{Fong2013}.

Preventing the widespread adoption and effectiveness of human-robot teams with natural language communication, there are at least two remaining concerns about the trustworthiness of such applications. First, systems for natural language processing (NLP) are trained on large text corpora with no direct sensory experience of the real world. Robots with NLP capabilities may know syntax and grammar, and may even be able to produce valid texts, but they may not understand how language relates to sensory-motor experiences in the real world. This problem is widely known as the \textit{language grounding} problem. Second, society at large is generally skeptical about the integration of automated systems, and users do not always \textit{trust} the robot. Hancock (2011) distinguishes between trust in intention and trust in competency \cite{Hancock2011}.  With trust in intention, the human believes that the robot is not deceptive and that it will aim for the goals of the team. With trust in competency or ability, the robot is believed to have the hardware and software needed to solve the task. Beyond cybersecurity and privacy concerns, trust in intention will typically be a reasonable assumption for most human-robot teams. By contrast, trust in capability is of major concern in human-robot teams.

This paper considers the interaction between the above two concerns; if language grounding is ``where robotics and NLP meet'', then we are interested in ``where HRI meets NLP''. We propose that language grounding comes with its own variant of trust, which we call \textit{trust in language grounding}; does the human trust the robot to understand language deeply, in the same sense that humans do, by understanding how all the objects, actions, and abstractions related to each other and how they might look or feel like in the real world? We investigate this topic by surveying the literature on language grounding (Section~\ref{sec: language grounding}), hypothesising key trust factors for trust in language grounding (Section~\ref{sec: trust factors}), and by conducting an empirical study to test these trust factors in a questionnaire and explore other trust factors in semi-structured interviews (Section~\ref{sec: case study}). The study is then concluded by proposing future research directions (Section~\ref{sec: future}) and summarising the findings (Section~\ref{sec: conclusion}).

\section{Language grounding}
\label{sec: language grounding}
The problem of language grounding is situated in a rich historical context but is of particular relevance today with increasingly powerful AI techniques and data sets. With the aim of discussing the key capabilities involved in language grounding, this section provides a historical background and reviews the various AI approaches and data sets available for understanding natural language (see Table~\ref{tab: languagegrounding AI}--\ref{tab: languagegrounding datasets} for an overview). 

\subsection{Historical context and definition}
In the early days of AI, reasoning about logical expressions formed from symbolic representations of real-world objects was thought to be the key approach to artificial intelligence. From the 80s onwards, this research agenda became less popular as researchers became aware that segmenting the real world into objects and understanding language are much more ambiguous and difficult than previously thought.

Consequently, researchers were puzzled with how an AI could potentially learn a language and capture its full relation to meaning in the full world.  With so much ambiguity in language, the autonomous association of meaning with verbal utterances and mathematical symbols seemed to be impossible within the traditional symbolic AI approach. Thus, the \textit{symbol grounding problem} \cite{Harnad1990}  was formulated:
\begin{quote}
How can the semantic interpretation of a formal symbol system be made intrinsic to the system, rather than just parasitic on the meanings in our heads? How can the meanings of the meaningless symbol tokens, manipulated solely on the basis of their (arbitrary) shapes, be grounded in anything but other meaningless symbols?
\end{quote}
As a potential solution, Harnad (1990) \cite{Harnad1990} proposed connectionism to detect the objects to which the symbols refer and then use a symbolic AI to compose syntactical relations. To understand this argument,  one must consider the framework of semiotics (see e.g. \cite{Short2007}), which interprets language in terms of 1) objects, which are entities observed in the real-world (e.g., a particular book located on your shelf); 2) concepts, abstractions formed from observing different instances of comparable objects (e.g., the notion of ``a book''); 3) a method to determine whether or not a particular object belongs to a particular concept (e.g., can we consider the item on the shelf to be a book?); and 4) a symbol, which is a label that is used to represent a particular concept (e.g., the characters, verbal utterances, or diagrammatic representations to refer to the concept of book). 

In present days, hardware and algorithmic advances have shown capabilities of connectionist AI to perform speech recognition and learn many aspects of language from a vast sensorimotor dataset. With this in mind, many researchers believe that it is possible to capture the full richness of language, including the large semiotic networks it entails. Indeed, some researchers believe that the challenge of the symbol grounding problem is solved \cite{Steels2007}. Therefore, currently, researchers speak particularly about the \textit{language grounding problem} to refer to the remaining difficulties in developing the rich network of meanings to provide the robust language comprehension that is so characteristic of human adults. 

Language grounding research is based on different technologies such as NLP techniques that capture the meaning of large bodies of text and dialogue managers that perform effective dialogue. Such technologies should provide an understanding that is sufficient for all the aspects of the AI's task in the real world or even a full  human-level understanding of natural language. Humans thoroughly understand natural language in terms of real world referents and are a prime exemplar for language grounding. When asked a query about a piece of text or within a spoken dialogue, humans are able to \textbf{reason} effectively about the intentions of others, the causal chain of events in a story, and to imagine a corresponding scenario in the real world. They provide \textbf{symbol grounding and sensorimotor learning}, being able to relate natural language to objects, abstract concepts, or daily activities acquired through trial and error. They effectively cope with \textbf{ambiguity}, manage \textbf{dialogue}, and interpret \textbf{non-verbal communication} as a supplement to natural language.

\subsection{AI approaches}
A wide variety of AI approaches are relevant for language grounding. The following subsection provides a categorisation consisting of five approaches, including symbolic AI, probabilistic machine learning, neural networks, reinforcement learning, and developmental robotics.
\paragraph{Symbolic AI}
The symbolic AI (or logic AI) approach is rooted in the works of early computer scientists such as G{\"o}del and Church, who devised logical calculus methods to compile executable programs from numerical expressions. The approach has been continued in terms of inductive logic programming \cite{Cropper2022} and cognitive architectures such as Soar \cite{Laird2012} and Cyc \cite{Lenat1990}, in which logical predicates are used to form automated reasoning systems. For language grounding, the symbolic AI is mainly pursued in terms of semantic parsing \cite{Kamath2018}, which transforms natural language onto a formal representation, such as first-order logic \cite{Becerra-Bonache2015} or lambda calculus \cite{Zettlemoyer2009,Thomason2015,Liang2016,Ross2018}. To obtain a deeper understanding of the text, most such semantic parsers are based on supervised learning over a large labeled dataset of
natural language utterances paired with annotated logical forms. To limit the amount of data required, one may also use weakly supervised learning based on partially labeled data sets and dealing with ambiguity is an important problem in this setting. Techniques have been developed to deal with homonyms or noise (e.g. \cite{Siskind1996}) but also for full sentence ambiguity; for example, one may use an incremental first-order logic algorithm that does not require explicit meanings of phrases but only requires a context, defined as a set of ground facts \cite{Becerra-Bonache2015}.

Despite their precision and efficiency in many problems, language grounding is a particular challenge for symbolic AI. If the meaning of objects is pre-programmed or derives from a set of ground facts then a symbolic AI does not really understand the meaning in terms of its real-world referents and neither does it connect these symbols to the rich semiotic network that we see in human adults; the link to sensorimotor learning is often completely overlooked. Approaches that do try to incorporate perception (e.g. \cite{Daoutis2009}) use pre-defined objects envisioned a prior by the designer, so the approach does not fare well in a priori unknown environments. 

Due to the above-mentioned reasons, symbolic AI is not popular in recent language grounding research. However, recent works in neuro-symbolic AI combine the strengths of neural networks with symbolic reasoning and logic programming for NLP applications \cite{Mao2021,Mao2019,Zellers2021,Hamilton2022}. For example, Neuro-Symbolic Concept Learner \cite{Mao2019} uses a neural perception module to classify objects within a visual scene and a semantic parser to translate questions into executable logic programs. State-of-the-art in semantic parsing has also shifted to large language models based on neural networks \cite{Einolghozati2019}; this often falls outside the scope of symbolic AI as here the output of the parser is often natural language rather than logical programs \cite{Rongali2022}. As a further shift away from the traditional logic-based semantic parsers, new representations for semantic parsing include labeled graphs which abstract away the meaning from different utterances  into a single natural language concept \cite{Banarescu2013} and hierarchical representations \cite{Lu2008,Gupta2018a}. A primary benefit of hierarchical representations is their ease for the user to understand task-oriented dialogue \cite{Einolghozati2019,Gupta2018a}.

\paragraph{Probabilistic machine learning}
In probabilistic machine learning \cite{Ghahramani2015}, one learns probabilistic models about the world from experience or from simulations. Probabilistic machine learning methods range from probabilistic programming (e.g. Monte Carlo sampling and hidden Markov models)
\cite{VandeMeent2018}, Bayesian optimisation \cite{Brochu2010}, data compression \cite{Shannon1948}, and automatic discovery of interpretable models \cite{Lloyd2014}. 

In language grounding research, probabilistic machine learning techniques allow to model the probability of true state of hidden variables in the environment (e.g. the presence of objects). In this context, Hidden Markov Models (HMMs), which describe a Markov process with hidden variables being predictable based on the current state of observable variables, can be used to disambiguate possible interpretations of words in ambiguous perceptual contexts based on video data \cite{Yu2013a}. Using the same principle, HMMs can also be used for eye fixation finding and clustering temporal sequences of human motions, where the clusters represent word meanings that can be associated with natural language using an expectation maximisation algorithm \cite{Yu2003}. Similarly, Bayesian filters, which model hidden variables based on a sequence of states of observable variables, have been used for probabilistic inference of object presence; for example, when the user asks to retrieve a particular object, the Bayesian filter will output the probabilities that objects are the one referred to in the instruction and this output can be used to determine whether to grasp an object or to further elaborate the question \cite{Zhang2021b}. 

Another relevant application of probabilistic machine learning is to provide a dialogue manager. A traditional approach is probabilistic planning \cite{SantosTeixeira2022}, in which a dialogue manager, equipped with a model of probabilistic state-transition dynamics, must plan consecutive actions from a starting state to a particular goal state. An alternative approach is to combine probabilistic reasoning from Bayesian networks with logic AI. For example, probabilistic rules may be formulated over state variables, which may be used for utility-based action selection \cite{Lison2016}.

Despite their benefits in accounting for uncertainty, the disadvantages of probabilistic machine learning are that (a) many if not all of the concepts, variables, and models have been pre-programmed by the designer; (b) the approach often has difficulty with coping with state and action spaces that are continuous or high-dimensional; and (c) modelling statistical dependence does not provide a timeline of how the variables develop and can often be due to hidden confounding factors that may not always be identifiable.

\paragraph{Neural networks}
Inspired by neurophysiological observations, artificial neural networks are simplified models of biological neurons as non-linear functions of  weighted sums of inputs. The weights represent the strength of connections between neurons and the non-linear ``activation function'' represents the threshold for the action potential to signal the activation of a neuron. 

Early ``symbol grounding transfer'' studies \cite{Cangelosi2006,Riga2004,Cangelosi2000} demonstrated how new concepts can be learned using neural networks. In one study, for example, a hybrid of supervised learning (Multi-Layer Perceptron, or MLP) and unsupervised learning (Self-Organising Map, or SOM) is trained to acquire new symbols through the input of linguistic combinations of previously acquired basic words (e.g. zebra as a combination of horse and stripes). The system is then tested by presenting the sensory information corresponding to the new concept (e.g. an image of a zebra) and observing the category matches. A follow-up study also showed that a similar grounding transfer could be achieved for action concepts, namely by presenting two action instructions allow the composition of actions into higher-order actions \cite{Cangelosi2006}. Such symbol grounding studies make a convincing philosophical point; however, learning a large body of concepts incrementally in this way is unlikely to be feasible in practice.

At present, attentional processes as well as feedback connections are being investigated in attentional neural networks and recurrent neural networks, both of which have become the industry standard in NLP (e.g. Google Translate). Recurrent neural networks are distinguished from their feedforward counterparts by the ability to model temporal dependencies, allowing to represent dynamical systems such as the weather, economic trends, and natural language. Natural language is one particular kind of dynamical system: the temporal dependencies in the sequence of vowels and consonants determine which word is formed and the sequence of words determine how to form a sentence. Long Short-Term Memory networks (LSTMs) \cite{Hochreiter1997} use self-loops and gating mechanisms to ameliorate the vanishing gradient problem, allowing to learn much longer sequences than traditional recurrent neural networks. Attentional networks such as Transformers and related methodologies  \cite{Vaswani2017,Kitaev2020} can often outcompete recurrent neural networks.  Transformers \cite{Vaswani2017} make use of attention heads, which are query, key, and value matrices to interrelate consequent words. Different such attention heads can provide different definitions of relevance. For example, one head may specialise in  pronouns and possessives linked to an earlier mentioned person or object whereas another would link the subject, verb and noun together; their combination can then be used to parse the sentence ``Anna called him. She wanted to tell him that she is sorry''. 

A critical precondition for the success of neural networks in NLP is how the words in a corpus of data are embedded into vector representation.  Word2vec \cite{Mikolov2013} was one of the first such embeddings and was based on a shallow 2-layer neural network. Global Vectors (GloVe) similarly provides a unique vector representation for each word but is based on unsupervised learning of the co-occurrence of words \cite{Brennan2017}.  Recent works realised that Word2vec lacks context and have therefore investigated \textit{contextual} embeddings. For example, Embeddings from Language Models (ELMo) \cite{Peters2018} uses not only parses the current word but also parses its context -- i.e. the left and right part around the word of interest -- through the same deep neural network. What is now one of the most advanced models in this sphere is Bidirectional Encoder Representations from Transformers (BERT) \cite{Devlin2019}. BERT is bidirectionally contextual, in the sense that it parses from left to right and from right to left jointly in all layers. It applies deep unsupervised pre-training to provide the contextual representation within the target neural network translation architecture. Contextual representations have also been investigated in multi-lingual contexts. For example, BERT has been trained on different monolingual corpora to create multilingual representations, which have been shown to yield zero-shot transfer to new languages with zero lexical overlap \cite{Pires2020}. Related techniques such as ALBERT \cite{Lan2020}, an adapted version of BERT with lower memory and computation requirements. While BERT-related text-encoding techniques are based on masked language generation, replacing some tokens with a mask and then formulating a loss based on how well the unmasked input token was reconstructed, another approach is to train the text encoder as a discriminator, that is, to train the text-encoder on corrupted tokens and assessing whether or not a token is corrupted or not; this approach is taken in ELECTRA \cite{Clark2020}, which outperforms BERT.

 In language grounding research, neural networks typically provide translation, speech recognition and/or predictions as a supplement to other AI components systems (e.g. reinforcement learning). Provided that neural networks have a suitable training and integration with other components, they have been applied to grounding spatio-temporal concepts such as ``to the left of the lamp'' and the distinction between  ``is'' and ``was'' \cite{Karch2021}, and visual grounding \cite{Yu2016,Deng2021,Zhang2021b}, in which object comprehension and the localisation of the referents are the main topics of ongoing research. The attention mechanism of transformers, for example, can be applied to different streams of inputs, the visual and the language input \cite{Deng2021}, allowing to effectively apply selective attention to particular parts of the images and text and correlate them with each other. To predict the correct description of an object within visual scenes, one may devise a system which uses the sequential processing of LSTMs in conjunction with convolutional neural networks, and one may distinguish between different such objects within a scene e.g. the blue ball vs the red ball by coupling different such LSTMs together \cite{Yu2016}.

Despite the impressive performance on complex language corpora and translation tasks, using neural networks comes with two disadvantages. First, neural networks have low transparency, where the decision-making process is not easily explained to the user. Second, neural networks require vast data sets and generalisation from limited data is a difficult challenge.

\paragraph{Reinforcement learning}
Reinforcement learning (RL) is a branch of machine learning with roots in psychological and physiological work on conditioning -- e.g. Pavlov's work on dogs' conditioned reflexes \cite{Pavlov1927} and the work by Skinner on operant conditioning of rats to learn to pull a lever for food \cite{Skinner1938}. Reinforcement learning is a paradigm which an agent interacts with an environment by perceiving the environment state or an incomplete observation, performing an action, and receiving a real-valued reward which indicates how well it performs. The agent can then learn a behaviour policy that optimises a long-term function of the reward.

In language grounding contexts, one common use of RL is to perform a traditional RL task, such as maze navigation and robot arm control, while being guided by natural language instructions. In such language-conditioned policies, a recurrent network embedding of a language-based instruction is merged with the agent's state to form the input of the policy \cite{Hermann2017,Chan2019} or the reward function \cite{Bahdanau2019}.  While the approach is common, unfortunately it has limited biological plausibility: since language becomes a pre-requisite for sensorimotor learning, which violates the decoupling of object perception, goal-oriented behaviours, and language observed in children. Moreover, an instruction in a given context only generates a low diversity of behaviors (i.e. a unique behavior for a deterministic policy or minor noise-induced behavioral variations for a stochastic policy). Addressing these two limitations,  Colas et al. (2020) \cite{Colas2020a} use language to generate a diversity of goals by first supplying an instruction, transforming it to an embedding (e.g. word-vector), and then repeatedly calling a goal-generator with different random seeds. 

Another relevant use of RL is as a dialogue manager, which is trained on a user simulation model or on real user data to output an utterance as an action to maximise a long-term utility function based on rewards. For interpretability and analysis, such utterances may be classified into abstract categories such as informing the user, selecting between two values (clarification), yes-or-no confirmation, and open-ended requests for more info from the user \cite{M.GasicM.HendersonB.Thomson2012}.  Sample efficiency is a key concern for RL-based dialogue managers, which has been improved by using Gaussian processes \cite{M.GasicM.HendersonB.Thomson2012,Gasic2014} and Kalman filters \cite{Pietquin2011} for value function approximation. Another approach is Interactive RL, where one uses action-specific feedback in addition to the task-specific reward to achieve faster convergence, which can be used to learn task-oriented dialogue managers that rapidly adapt to user preferences unseen during training \cite{Shah2016}. Ambiguity is accounted for within methods that use the partially observable Markov decision process as a framework \cite{Roy2000,M.GasicM.HendersonB.Thomson2012,Gasic2014,Lison2013a}. Model-based reinforcement learning has been used for explicitly modelling the transition dynamics and their uncertainty \cite{Lison2013a}. Recently, there is also a shift towards the ambitious setting of open-domain dialogue in which the user can ask literally anything and conduct a conversation as long as desired, for which state-of-the-art approaches are hierarchical \cite{Saleh2020} and mixture-of-expert approaches \cite{Chow2022}. Adapting to user sentiment is also an important feature of recent RL systems  \cite{Shi2018,Chow2022}.

\paragraph{Developmental robotics}
Developmental robotics is a related framework for interactive learning that is inspired by philosophical and developmental psychology traditions which recognise the need to understand language and cognition in a developmental context, much like children have to form their thoughts and learn language. Some emphasise the importance of embodiment and interaction in the physical world, and state that language mainly serves as labels to capture the real-world concepts, similar to the above-mentioned decoupling argument by Colas et al. (2020) \cite{Colas2020a}. For example, Piaget \cite{Piaget1952} believed there to be different developmental phases, including the understanding that objects remain when hidden (sensorimotor phase), learning that symbols can point to objects in the real world (pre-operational phase), learning to perform mental operations to simulate events (concrete operational phase), and understanding and manipulation of formal, abstract concepts (formal operational phase). Others emphasise the role of the social context, and especially the role of dialogue with others, in developing thinking individuals. For example, Plato emphasises the importance of dialogue, and even conceives of thinking as a dialogue of the individual with itself \cite{Dixsaut1997}. Similarly, Lev Vygotsky, in his treatise on sociocultural theory on development, stressed that children learn their norms, values, reasoning, and indeed language itself, by conversations with experts \cite{Vygotsky1978}. The influence of the environment has been studied in developmental psychology as the nature versus nature debate, and this debate has also been prominent in the theory of language acquisition. Nativist theories suggest that language is mainly innate. For example, Chomsky hypothesised an in-built Language Acquisition Device, which hypothesises innate knowledge of grammar based on stimulus poverty, the argument that children are not exposed to rich enough data to learn full grammar,  and universal grammar, the argument that in most normal conditions children will develop particular types of grammatical rules (e.g. distinguishing nouns from verbs) \cite{Chomsky1965}. 

At present, many of the above ideas have been investigated further, including in the context of robotics and AI. In language acquisition, many points go in favour for \textit{nurture}:  beyond empirical evidence in language acquisition \cite{DeBot2015}, general-purpose machine learning methods such as recurrent neural networks have no innate knowledge of grammar but are able to perform human-level language translation and speech recognition. However, there is the point to be made for \textit{nature} that the human system as well as the recurrent neural networks have a favourable inductive bias for learning grammar, and some authors also describe an innate language acquisition device in terms of inductive bias \cite{Briscoe2000}. Despite the impressive gains of recurrent neural networks, language translation and speech recognition do not necessarily imply \textit{grounding}. To achieve grounding, developmental roboticists consider the wider context of learning in the following sense: \\
1) social context: for example, in shared language evolution \cite{Steels2007},  multiple robots interact and based on success on the task at hand converge on a shared lexicon (much like Vigotsky's reasoning) e.g. color or spatial concepts; \\
2) embodiment: by observing the effect of actions and sensory data in the real world, words become labels to existing sensorimotor concepts (much like Piaget's reasoning); and\\
3) rich multimodal datasets: see \cite{Heinrich2020} for a recent developmental approach for grounding with cross-modal fusion and adaptive time scales. \\

A variety of early approaches to the symbol grounding problem have been explored to transferring between learned models. Earlier work showed two possible routes, one via shared language evolution and one via neural networks. In shared language evolution, there is an interaction between different robots helps to transfer one robot's concept to another robot. Such works have been shown to create a shared lexicon, e.g. in terms of color \cite{Steels2007} and spatial concepts \cite{Spranger2015}, but a) while this addresses the symbol grounding problem, it does not address the full language grounding problem; and b) the approach does not work if the user only have one robot. It was also shown that semantics as well as syntax can be co-evolved from robot-robot interaction \cite{Spranger2015}, which brings a sound philosophical foundation to the role of nurture in language acquisition and language grounding, but unfortunately this approach would not necessarily yield human language, which would be essential for practical human-robot settings and especially so if trust is to be considered. However, it is also possible to draw on the social context of human-robot dialogue for language grounding; in Thomason et al. (2019) \cite{Thomason2019}, natural language is translated into actions for a robot to perform a pick-and-place task and in which clarifying dialogues are used to assist in symbol grounding and language parsing. The AI technique includes a semantic parser to translate user commands into semantic slots based on desired actions, perceptual properties, and the source and destination of the object to pick up, a language grounder which uses maps and concepts to identify matching candidates to the perceptual info, and a clarifying dialogue to remove ambiguity as well as improve the conceptual models on-the-fly.

\begin{landscape}
\begin{table}
\caption{An overview of AI approaches for language grounding. The annotation ``(generic)'' indicates that most instances of the approach have the mentioned capability.}\label{tab: languagegrounding AI}
\begin{tabular}{p{3.5cm} p{3cm} p{3cm} p{3cm} p{3cm} p{3cm}}
\toprule
					   & \textbf{Capabilities}  &            &      &  \\ 
					   \midrule
\textbf{AI Technique}  &  Reasoning   &  Symbol grounding \& Sensorimotor learning  & Ambiguity  & Dialogue & Non-verbal communication  \\
\hline
Symbolic AI			& semantic parsing to first-order logic \cite{Liang2016} or lambda calculus \cite{Zettlemoyer2009,Thomason2015,Liang2016,Ross2018} &  combination with other approach (e.g. neuro-symbolic AI  \cite{Mao2021,Mao2019,Zellers2021,Hamilton2022} and modern cognitive architectures \cite{Laird2012})  & ground facts as context  \cite{Becerra-Bonache2015}  & task-oriented dialogue \cite{Einolghozati2019,Gupta2018a,Thomason2019}  &  combination with other approach (e.g. neuro-symbolic AI  \cite{Mao2021,Mao2019,Zellers2021,Hamilton2022} and modern cognitive architectures \cite{Laird2012}) \\ 	
Probabilistic ML &  conditional independence and Bayesian graphs (generic)   &  learning from annotated videos \cite{Yu2013a} & Hidden Markov models \cite{Yu2013a}, Bayesian filters \cite{Zhang2021b}  & probabilistic planning \cite{SantosTeixeira2022}, Bayesian networks \cite{Lison2016} &  categorising human motions \cite{Yu2013a,Yu2003} \\  						
Neural networks  			&  Spatio-temporal reasoning \cite{Karch2021,Deng2021,Yu2016}, Compositional reasoning \cite{Hudson2018,Gao2020,Wu2021}    &  Visual grounding \cite{Yu2016,Deng2021,Zhang2021b}  & Contextual embeddings \cite{Peters2018,Devlin2019,Lan2020,Clark2020}  & Representing the conversation history \cite{Chow2022,Shi2018}  & sentiment analysis \cite{Zadeh2018} \\					
Reinforcement learning		&  spatio-temporal reasoning and planning for dialogue and task (generic)   &  Language grounded RL \cite{Hermann2017,Chan2019,Bahdanau2019,Colas2020a}, sensorimotor learning (generic)  & POMDP-based dialogue managers \cite{Roy2000,M.GasicM.HendersonB.Thomson2012,Gasic2014,Lison2013a} & MDP-based \cite{Pietquin2011,Chow2022,Shi2018,Peng2017,Shah2016} and POMDP-based \cite{Roy2000,M.GasicM.HendersonB.Thomson2012,Gasic2014,Lison2013a} dialogue managers &  sentiment analysis for dialogue managers \cite{Shi2018,Chow2022} \\						
Developmental Robotics		&  /   &  Multimodal grounding and embodiment \cite{Heinrich2018,Heinrich2020}, Social context \cite{Steels2007,Spranger2015,Thomason2019}  & Multimodal grounding and embodiment \cite{Heinrich2018,Heinrich2020} & Social context \cite{Steels2007,Spranger2015,Thomason2019} & Multimodal grounding and embodiment \cite{Heinrich2018,Heinrich2020}  \\
\bottomrule
\end{tabular}
\end{table}
\begin{table}
\caption{An overview of data sets for language grounding. The annotation ``(generic)'' indicates that most instances of the approach have the mentioned capability.}\label{tab: languagegrounding datasets}
\begin{tabular}{p{3.5cm} p{3cm} p{3cm} p{3cm} p{3cm} p{3cm}}
\toprule
					   & \textbf{Capabilities}  &            &      &  \\ 
					   \midrule
\textbf{Data} &  Reasoning   &  Symbol grounding \& Sensorimotor learning  & Ambiguity  & Dialogue & Non-verbal communication  \\
\hline
Data sets        &  Compositional reasoning \cite{Johnson2017,Lake2018,Andreas2020,Wu2021}, Causal reasoning \cite{Roemmele2011,Ponti2020,Du2022,Yang2022,Mariko2020,Dua2019}  & Visual grounding \cite{Kazemzadeh2014,Yu2016}, multimodal grounding and sentiment analysis \cite{Heinrich2018,Hasan2019,Poria2020} &  multimodal grounding and sentiment analysis \cite{Heinrich2018,Hasan2019,Poria2020}  & Task-oriented dialogue \cite{Einolghozati2019,Gupta2018a}  &  multimodal grounding and sentiment analysis \cite{Heinrich2018,Hasan2019,Poria2020} \\                                                
Simulation environments &  BabyAI \cite{Chevalier-Boisvert2019}, Minecraft \cite{Aluru2015}   & BabyAI \cite{Chevalier-Boisvert2019}, Minecraft \cite{Aluru2015}, DeepMind Lab \cite{Beattie2016}  & BabyAI \cite{Chevalier-Boisvert2019},Minecraft\cite{Aluru2015}, DeepMind Lab \cite{Beattie2016}  & High-fidelity simulators for task-oriented dialogue \cite{Beetz2018,Milliez2014} & ROS4HRI \cite{Mohamed2021} \\
User interfaces  &  /   &  Remote control, gestures, and haptic feedback \cite{Rouanet2011}  & Remote control and haptic feedback \cite{Rouanet2011}, bio-feedback \cite{Rani2005}  & GUIs \cite{Li2020a}, web interfaces  \cite{Thomason2015}, Mechanical Turk \cite{Thomason2015,Saleh2020}  & GUIs \cite{Li2020a}, Web interface \cite{Burgard1998,Thomason2015}, Remote control, gestures, and haptic feedback \cite{Rouanet2011}, bio-feedback \cite{Rani2005}  \\
\bottomrule
\end{tabular}
\end{table}
\end{landscape}

\subsection{Providing data to the AI}
\begin{figure}
\centering
\subfloat[Ref-COCO (credit: \cite{Yu2016})]{\includegraphics[width=0.35\textwidth]{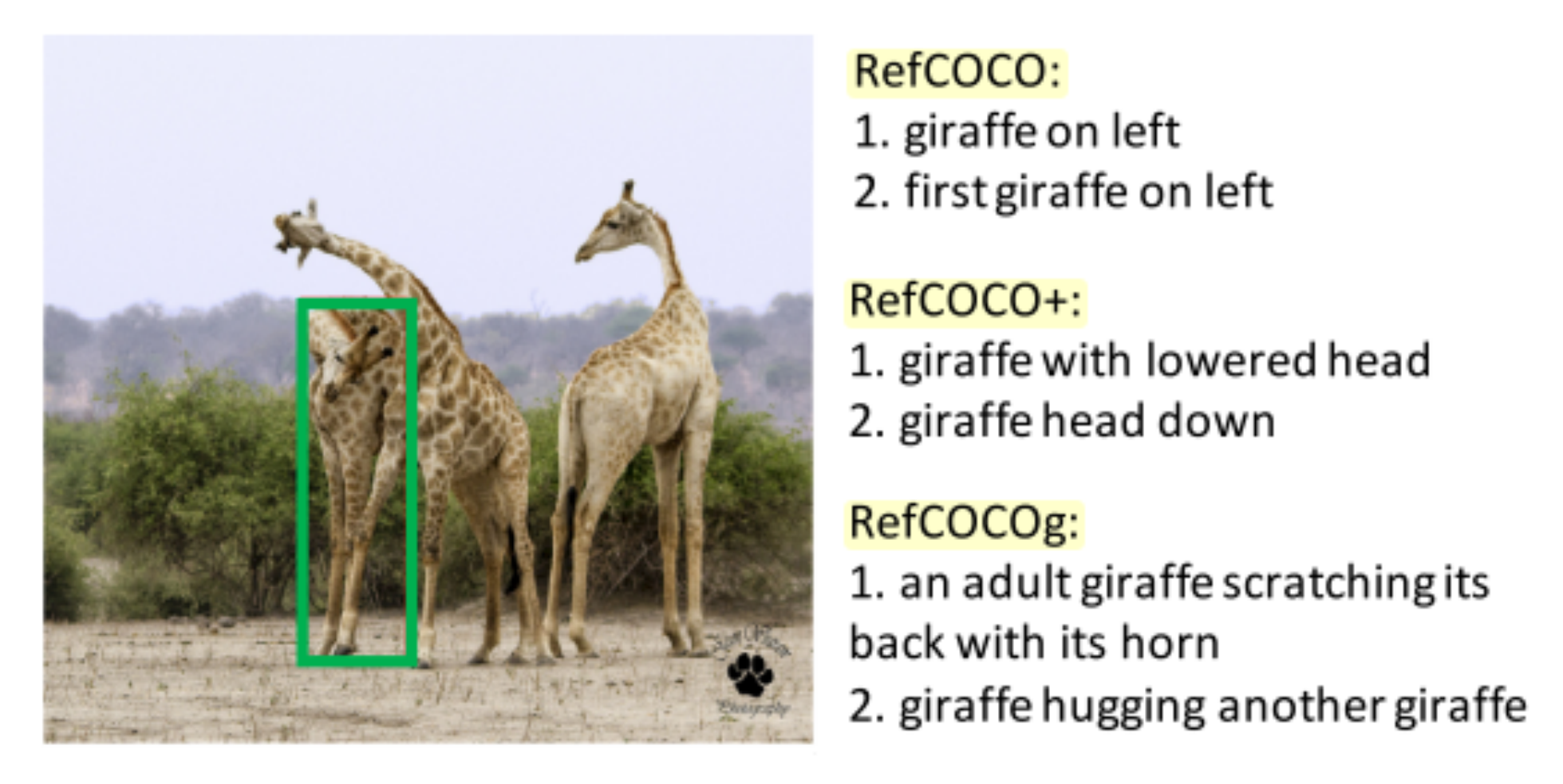}} 
\subfloat[Minecraft (credit: \cite{Aluru2015})]{\includegraphics[width=0.225\textwidth]{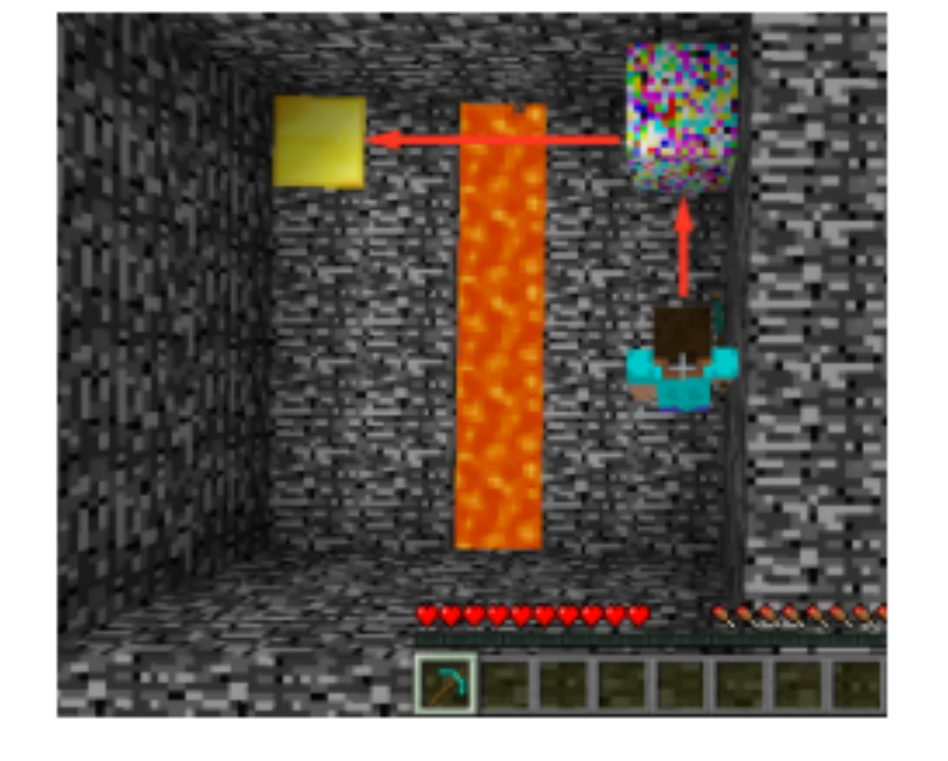}} 

\subfloat[User interfaces (credit: \cite{Li2020a})]{\includegraphics[width=0.225\linewidth]{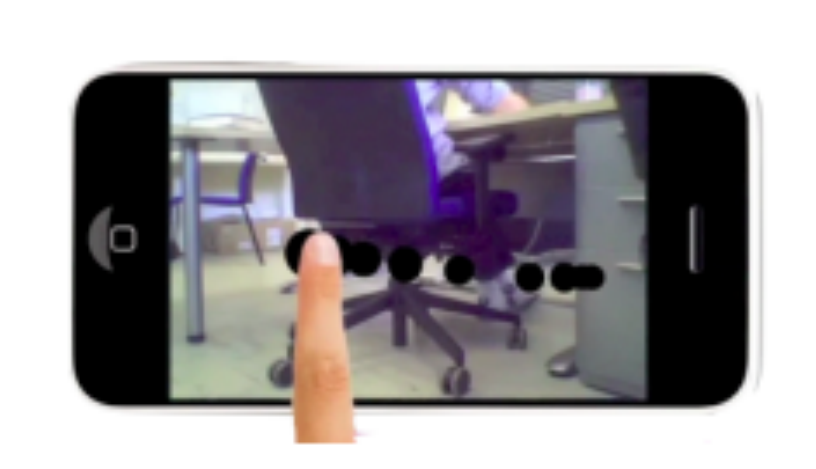}     \includegraphics[width=0.225\linewidth]{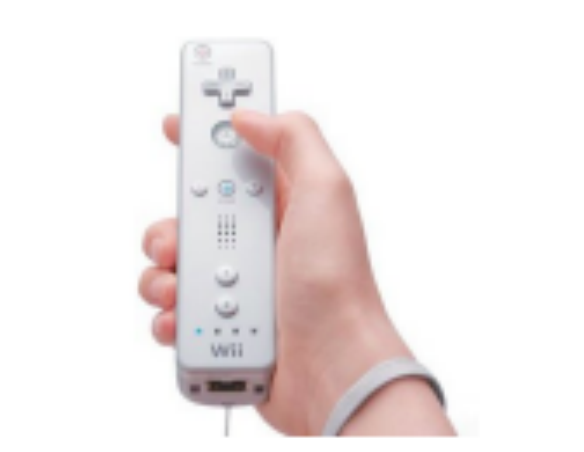} }
\caption{Providing data to the AI. \textbf{(a) Ref-COCO:} data sets such as Ref-COCO allow visual grounding of objects by coupling visual scenes with a verbal description; \textbf{(b) Minecraft:} interactive simulation environments such as Minecraft provide a rich environment for interpreting and executing language-based instructions; and \textbf{(c) User interfaces:} user interfaces allow clearly indicating the desired task and improving the AI by user feedback, with an iPhone GUI and Wii remote control being provided as examples.} \label{fig: data}
\end{figure}

Many of the promising AI approaches (neural networks, reinforcement learning, developmental robotics) require high-quality data to be able to parse subtle nuances in language. Such data may be available offline from recorded data sets but may also be gathered interactively online by learning in a simulation environment or by specially designed user interfaces (see Figure~\ref{fig: data}). While some of these were briefly mentioned above, this subsection provides a more extensive overview.

\paragraph{Data sets}
A variety of data sets have been designed specifically for combining language with sensorimotor data from various modalities. Visual grounding data sets are based on video or image data. Ref-COCO \cite{Kazemzadeh2014,Yu2016} is a large dataset for learning how to refer to a particular object in an image using a text sentence. Ref-COCO covers 80 common objects and for each of these has a large number of image-text combinations. The top performer on Ref-COCO and related data sets is TransVG (see \cite{Deng2021} for a comparison). Full video and audio streams are also found, particularly in the field of sentiment analysis (see \cite{Poria2020} for an overview), where the challenge is to understand the intent and attitude of a person based on non-verbal cues such as prosody, gestures, and facial expressions. For example, understanding of humour can be studied using the UR-FUNNY data set \cite{Hasan2019}. One of the top performers on the UR-FUNNY data set as well as other sentiment analysis data sets is the Memory Fusion Network \cite{Zadeh2018}, which combines memory with LSTMs for each modality with an Transformer-based attention layer and a multi-view gated memory.

 Other multimodal datasets are specific to a particular robot platform, which is beneficial for full language grounding. For example, for the Neuro-Inspired COmpanion (NICO) humanoid robot, the EMIL dataset \cite{Heinrich2018}  collects auditory, sensorimotor, and visual modalities with a verbal description for grounding words into physical meanings. The verbal description can take the form of phonetic data or word embeddings (e.g., Word2Vec , GLoVE \cite{Brennan2017}). As another example, Matuszek et al. (2014) \cite{Matuszek2014} investigate gesture recognition in addition to other modalities on a Gambit manipulator robot arm. They constructed an RGB-D video and speech dataset to allow a robot to identify the objects pointed to by language. A total of 234 scenes were collected, each involving two or more objects to be indicated. Language from the corpus collected was hand-transcribed although the robot experiments used speech recognition. Data was collected from 13 participants, each of which described 28 scenes, yielding 364 language/video pairs. Such platform-specific data sets unfortunately make it difficult to assess a particular algorithm for another robot platform; therefore, we refer to the algorithms in the respective papers for the most appropriate AI technique.

Visual grounding can also be coupled with compositionality, the ability to analyse sentences in terms of constituent expressions and to arbitrarily be able compose and reason about such expressions in different ways. Compositional reasoning in this sense can be studied using the CLEVR \cite{Johnson2017}, SCAN \cite{Lake2018}, gSCAN \cite{Andreas2020}, and ReaSCAN \cite{Wu2021} data sets. These data sets involve a large number of visual scenes, each containing different combinations of objects and questions, probing the ability to reason about the relationship between the various objects (e.g. location, color, shape, material) in the scene. Among the best-performing algorithms for gSCAN and ReaSCAN, two of the most challenging benchmarks assessing systematic generalisation and compositional reasoning, is the graph convolutional neural network with attention-based bidirectional LSTM layers as proposed in Hudson et al. (2018) \cite{Hudson2018,Gao2020,Wu2021}.

Tests for causal reasoning over text corpuses have also been implemented, primarily in question-answer format, and among the best performers one typically finds variations of BERT or GPT-3. COPA \cite{Roemmele2011} is a list of premises each of which is followed by a two-choice answer which provides a valid implication. The X-COPA data set \cite{Ponti2020} is a multilingual version of COPA, featuring 11 languages. The e-CARe data set \cite{Du2022} is similar to COPA but provides causal explanations by means of human annotations and, with its size of 21,000 multiple-choice causal explanation questions, it is one of the largest human-annotated commonsense causal reasoning dataset available. FCR \cite{Yang2022} provides an even larger data set, with 25,193 cause-effect pairs as well as 24,486 question-answering pairs, which ask ``why'' and ``what-if''  questions. The data set also tests for event extraction and span-based QA which includes three different fine-grained causalities, namely causing, enabling and preventing an effect. While event extraction and span-based QA have been addressed together in FCR, there are also data sets available that focus on these aspects exclusively (see \cite{Yang2022} for an overview). For example, causality extraction is featured in FinCausal 2020 \cite{Mariko2020}, where one refers to text chunks as causal events, and DROP  \cite{Dua2019} requires the addition, counting, or sorting of particular references within a paragraph. 

Finally, a wide variety of NLP data sets, some of which are relevant for language grounding, can be found in the yearly SemEval workshop's data sets (see \url{https://semeval.github.io/}). Typically, the highest performers within SemEval use state-of-the-art neural network techniques for NLP together with some domain-specific adjustments. For example, in SemEval 2021, one data set involved sentence classification, phrase recognition, and triple extraction to automatically structure the contributions of publications, and Liu et al. (2021) \cite{Liu2021} received the best system award for their system that best solved this task using a combination of BERT-based classification and sequence labeling models with rule-based methods. As another notable example from SemEval 2021, the Reading Comprehension of Abstract Meaning task was best solved by a AI technology combining task-adaptive pretraining with multi-head attention (with Transformers) \cite{Zhang2021}.

\paragraph{Interactive simulation environments}
In reinforcement learning approaches, it is common to allow the robot to learn from experience in a simulated, interactive environment. We distinguish here between game-like environments and high-fidelity robotic simulations.

Game-like environments have relatively low realism, and therefore may not transfer well to real-world human-robot teams, but are of conceptual interest in learning to solve challenging tasks based on language-based instructions. DeepMind Lab (DML) \cite{Beattie2016} provides a rich 3D simulation platform providing large and complex environments with rich visuals, partial observability, and the possibility to formulate a wide variety of tasks. Two studies on language grounding were conducted in a DML environment, showing that after a language-based instruction at the start of the episode, RL agents can successfully find the correct item that was instructed \cite{Hermann2017,Hill2020}. Minecraft provides a similarly rich environment and several works have explored how to learn a task based on language instructions (e.g. \cite{Aluru2015}). While similar to DML, currently the work has focused on relatively simple tasks such as finding a particular object, one of the key benefits of the rich Minecraft environment is that it allows to consider how to perform more abstract or hierarchical tasks such as ``build a house''. The BabyAI environment \cite{Chevalier-Boisvert2019} is particularly relevant for including humans into the loop of grounded language learning. It includes different levels, in which an agent receives a particular instruction in a subset of English and must pick up and drop off objects within a partially observable gridworld. The levels are incremental, in the sense that higher levels require new skills as well as the skills needed in the previous levels and the complexity of the language instruction grows over levels. The environment includes humans in the loop by allowing them to actively select demonstrations to learn from for improved sample efficiency.

High-fidelity simulation engines such as ROS, MORSE, and Gazebo have also been useful for HRI studies in language grounding. A typical use, primarily found in the context of task-oriented dialogue \cite{Beetz2018,Milliez2014}, is to provide realistic robot motions, repeatable experiments, and the ability to systematically investigate labelled features of the robot. Another use is to provide social signal processing \cite{Pantic2014} to the robot; for example, the  ROS4HRI \cite{Mohamed2021} provides a framework that provides labelled features and IDs about the human (e.g. demographic factors, gaze, expression, height, voice) with which the robot is interacting.

\paragraph{User interfaces}
User interfaces can facilitate or supplement language-based communication between human and robot through monitoring, remote assistance, graphical illustration, etc. Burgard et al. (1998) \cite{Burgard1998} presented a web interface for a museum robot where the users can remotely follow a guided tour. The web interface includes a bird's eye view, the view from the robot's camera, and a view of the exhibition. Rouanet et al. (2011) \cite{Rouanet2011} study 4 different interfaces designed to teach new visual objects to a Nao social robot, including the following:
\begin{enumerate}
\item iPhone GUI: a video stream of the robot camera is displayed on the screen, allowing users to monitor what the robot sees, and the touch screen allows user to point towards particular objects or to swipe from one location to another to suggest particular motions to the robot.;
\item Wiimote interface: the remote control of the Wii is used to direct to a part of the robot and then using the directional cross one can instruct the robot to move;
\item Wiimote with laser pointer: laser points to an object, and the robot gets positive haptic feedback when the user rumbles the Wiimote;
\item Gestures: user makes any gesture and the robot interprets, however, note that Wizard-of-Oz (i.e. a human pretending to be a robot) was used for control commands because gesture recognition is (or was at the time?) still challenging.
\end{enumerate}
Similarly, more sensitive haptic interfaces also exist, which can be combined with a motion editor, with applications such as live performance art \cite{Lee2014}. SUGILITE \cite{Li2020a} is a GUI-based framework for 1) allowing the user to teach new tasks via verbal instruction and GUI-based demonstration; 2)  clarification of the spoken intent using GUI-grounded verbal instructions to clarify the user's intent; (3) providing task parameters, for example, by pointing to an image of an iced coffee and its language description as ``iced cappuccino'' to indicate what kind of drink the user wants for a Starbucks order; and 4) generalisation, applying taught concepts to different contexts and task domains. Mechanical Turk, an Amazon-based crowdsourcing service is another popular interface with the unique benefit of being able to gather data from a large number of users. For example, Thomason et al. (2015)  \cite{Thomason2015} used Mechanical Turk in their web interface to allow training a robot on user-requested office tasks based on batches of dialogues with hundreds of users to improve user satisfaction in a test batch of unseen users. Beyond such task-oriented dialogues, Mechanical Turk has also been used for open-domain dialogue \cite{Saleh2020}. Bio-feedback devices may be used to communicate the human's psychological state (e.g. anxiety) to the robot, which in turn may take actions to keep the human in a positive state and be more collaborative \cite{Rani2005}.

\section{Hypothesised trust factors for language grounding}
\label{sec: trust factors}
Trust factors in human-robot teams are typically subdivided into three categories following the categorisation of Khavas (2021) \cite{Khavas2021}. First, robot-related factors describe aspects of the robot appearance and behaviour or empathy that affect trust. Second, human-related factors describe aspects of the human such as prior experience, beliefs and expectations around robots, etc. Third, task-related factors such as the risk and safety and the nature and location of the task. We hypothesise that each of these categories will also be important to trust in language grounding but given the large number of factors we do not hypothesise these more specifically; instead, we will empirically test their importance in open-ended questions and semi-structured interviews of Section~\ref{sec: questionnaire}.

While the above factors have been studied extensively, we mainly focus on the language capabilities of the robot as these are specific to language grounding. We select such language capabilities based on the relevance to aspects of language grounding studied in our review, namely reasoning, symbol grounding, coping withg ambiguity, managing dialogue, and interpreting non-verbal communication. In particular, from these we select capabilities that can be probed with relatively straightforward questions such that a relatively broad audience can faithfully answer the question (see questionnaire in next section). We list the hypothesised factors below, along with a brief motivation:
\begin{itemize}
\item \textbf{C1: Clarifying ambiguity.} A variety of NLP methods are designed specifically for handling ambiguous, partially observable state space (e.g. \cite{Yu2013a,Zhang2021b}). Given the importance in understanding complex worlds, we hypothesise users to recognise observable attempts to clarify ambiguity through dialogue is important for trust in language grounding. 
\item \textbf{C2: Causal reasoning.} Causality can be understood from a conversational perspective: \textit{why} would the human such a question? Hilton's conversational model of causality \cite{Hilton1990} states that explanations  seek to identify the crucial difference between the target case (the question asked) and the counterfactual (another, related question that could have been asked) and that they are distinct from mere diagnoses in their relevance to the conversation. We hypothesise that users would think it is important for trust in language grounding for the robot to demonstrate, through dialogue, it understands why the question was asked.
\item \textbf{C3: Understanding task achievability.}  Planning turns out to be an essential element of many language grounding systems (e.g. \cite{Zhang2021b,Coradeschi2013}). We hypothesise that users it would be essential for trust in language grounding for the robot to demonstrate the ability to understand the task and discuss any questions about task achievement.
\item \textbf{C4: Temporal reasoning.} Temporal constraints are often desirable. Without awareness of time and temporal reasoning, the robot may fail to succeed in such tasks. We hypothesise that users believe temporal reasoning is important for trust in language grounding.
\item \textbf{C5: Gestures and intention.} The ability to infer the intention of the human from the instruction, and specifically which object or place is being referred to within the task, is an important requirement for task achievement. Since this depends not only on the instruction but also context factors, e.g. gestures and the visibility of the object or place, we hypothesise that users believe the ability to use gestures to infer intention is important for their trust in language grounding.
\item \textbf{C6: Learning from user feedback.} Machine learning methods such as reinforcement learning and neural networks have been used extensively for grounded language learning but this has been mostly in the form of ``objective'' instructions that do not take into account language preferences of the user. We hypothesise that users believe learning from user feedback is important for trust in language grounding.
\end{itemize}

\section{A case study with a human-robot cleaning team }
\label{sec: case study}
Having hypothesised various trust factors, we now turn to an empirical case study on a human-robot cleaning team, which consists of one human operator and a HELIOS UVC cleaning robot. Section~\ref{sec: experts} investigates expert opinions into what they think would be required for language grounding and trust in language grounding in the case study.
\subsection{The cleaning team}
\label{sec: cleaningteam}
The cleaning robot used in the case study is the HELIOS UVC robot, a disinfection and sterilization robot. The robot kills germs in the environment by decomposing their DNA structures, thus preventing and reducing the spread of viruses, bacteria and other harmful micro-organisms. The robot can move around and emit condensed UV beams to eliminate harmful micro-organisms in a 360 degree range around itself and has obstacle-avoidance sensors to avoid any collisions with nearby objects. In addition to these built-in capabilities, our case study assumes that the robot is trained to understand language and communicate with humans to help understand the task it is supposed to perform. Moreover, to ground language in visual observations of the environment, the robot is augmented with a camera and visual pattern recognition.

The human operator can provide speech-based instructions to the robot but may also communicate by other means, including gestures and visualisations. A representative cleaning team was captured in illustrative videos made within the CoBot maker space\footnote{See \url{https://cobotmakerspace.org/} for more details.}, a complex workshop environment with many obstacles, office tools, and even some other robots. 
\begin{figure}[htbp!]
\centering
\includegraphics[width=0.5\textwidth]{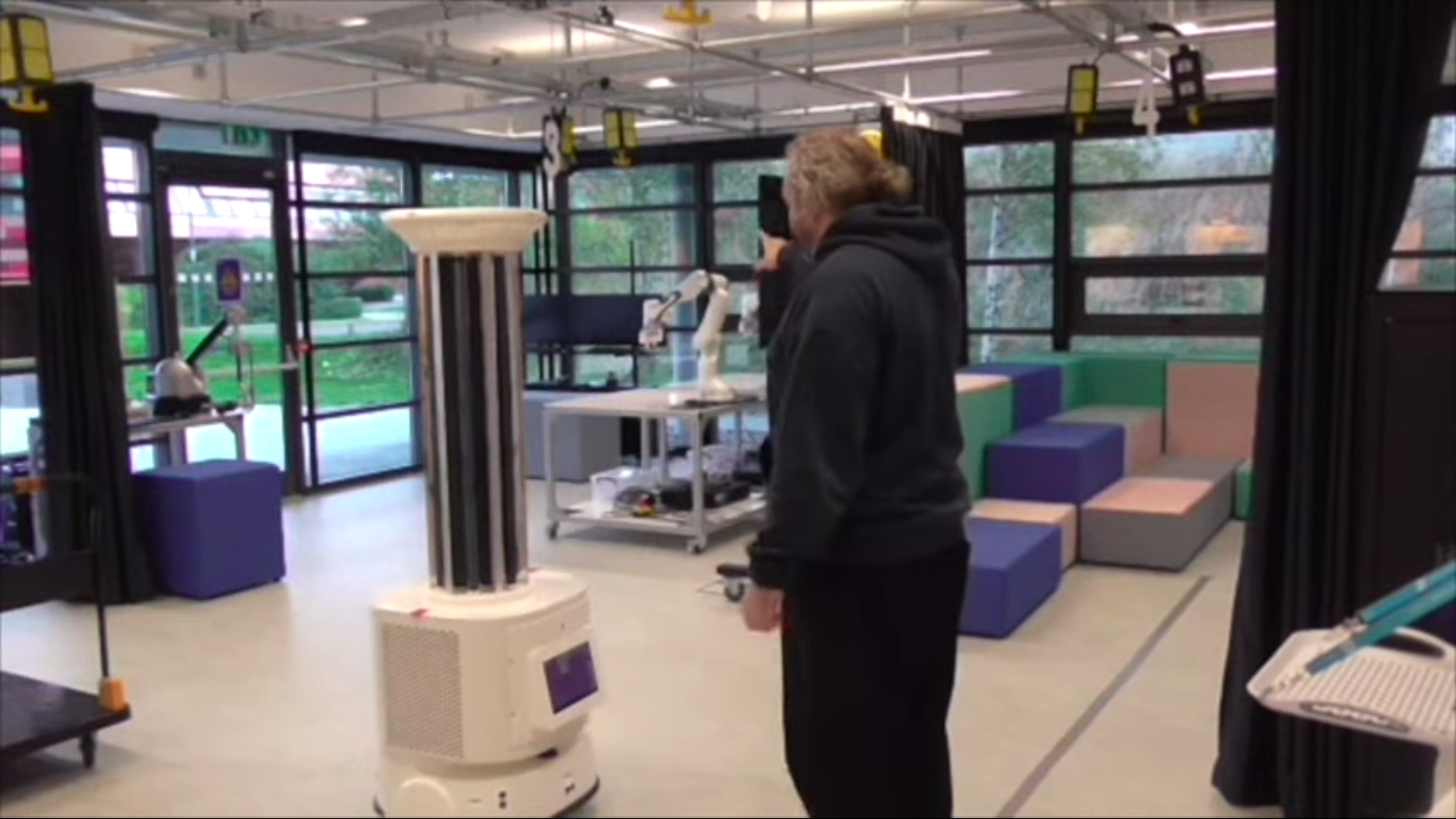}
\caption{The Helios UVC Robot and the human operator communicating with each other via natural language and gestures in a complex workshop with many obstacles.}
\end{figure}

\subsection{Questionnaire for the general public}
 \label{sec: questionnaire}
The questionnaire was composed of two parts. In a first part, the participants were asked multiple-choice questions probing to what extent a language capability affects their trust in the language grounding of the robot. In the second part, the participants were asked an open-ended question, where they could indicate anything else that would affect trust in the language grounding of the robot.
\paragraph{Multiple-choice questions} A total of six multiple-choice questions were asked. Five of these questions were supplemented with a video recording (see videos in Supplementary Materials) to illustrate the meaning of the question with a dialogue between the robot and human operator. Each question first explained the language capability, with or without an illustrating video, and then asked the participant ``How important is this capability for you to trust that the robot understands language as good as or better than a human?''. Response format for the open-ended questions was a 1 to 5 scale (Extremely important to Extremely unimportant). The six language capabilities  in Section~\ref{sec: trust factors} are operationalised as follows in the case study:
\begin{itemize}
\item \textbf{Clarifying ambiguity.} A video is shown in which a human operator asks the robot to clean the hallway. As there is more than one hallway, the robot then asks for a bit more clarification by first going in the direction of the nearest hallway and then asking whether the human means this hallway. After the  video, the participants are explained, ``This video illustrates a robot's capability to clarify ambiguous instructions.''
\item \textbf{Causal reasoning.} A video is shown in which a human operator asks to clean the area around the desk. The cleaning robot then asks whether it should also clean the area around the printer and the human replies yes. After the video, the participants are explained, ``This video illustrates the capability to reason about the ways in which humans use areas in a room. In this case, the robot recognises that people who use the desk are also likely to use the area around the printer. Therefore, when instructed to clean the area around the desk, the robot also offers to clean the area around the printer.''
\item \textbf{Understanding task achievability.}  A video is shown in which a human operator asks to clean the back of the room but the robot notices an obstacle and asks the human to remove it so that it can go to the back of the room. Participants are then explained, ``This video illustrates the capability to analyse a scenario and predict if there any obstacles that would prevent the robot from successfully completing the task.''
\item \textbf{Temporal reasoning.} A video is shown in which a human operator asks the robot to clean the room but stop before 5 PM. The partipants are then explained, ``This video illustrates the capability to reason about instructions that depend on time.''
\item \textbf{Gestures and intention.} A video is shown in which a human operator asks the robot to ``clean over there'' while pointing to the other side of the room. The participants are then explained, ``This video illustrates the capability to correlate language and intent with physical actions.''
\item \textbf{Learning from feedback.} The participants are explained, ``To help the robot adapt, the robot could be trained by user feedback to learn the intent of instructions. For example, the user may say `good' whenever the robot performs the task perfectly and `bad' whenever the robot has misunderstood the task.''
\end{itemize}
 
The responses to the multiple-choice questions (see Table~\ref{tab: capabilities}) show that most participants believe these capabilities to be either important (40--50\%) or extremely important (36--50\%), and all the capabilities are significantly higher-rated than the neutral score based on one-sided one-sample t-tests (all $p < 0.05$). Clarifing ambiguity and understanding task requirements have the highest percentage of ``extremely important'' ratings and they have the highest overall importance score (3.4). None of the participants rated any of the capabilities ``extremely unimportant'' and ``unimportant'' was selected less than 5\% for any of the capabilities.

\begin{table}
\caption{Effect of language capabilities on trust in language grounding. For each language capability, the corresponding importance is shown based on the response frequencies and the mean response. SE denotes the standard error as a measure of spread in responses. Overall importance score is computed as average importance rating across all participants.} \label{tab: capabilities}
\resizebox{\textwidth}{!}{
\begin{tabular}{l | P{2.5cm} P{2.5cm}  P{2.5cm} P{2.5cm}  P{2.5cm} P{2.5cm}}
\toprule
\textbf{Capability} & \multicolumn{5}{l}{\textbf{Frequency} ($\%$)} & \textbf{Overal importance} (Mean $\pm$ SE) \\
\midrule \\
					    & Extremely important (4) & Important (3) & Neutral (2) & Not important (1) & Extremely  unimportant (0) &  \\ \hline
Clarifying ambiguity	& $50.3$ 	& $43.0$ 	& $5.4$ 	& $1.3$ 	& $0.0$ 	& $3.4 \pm 0.36$ \\	
Causal reasoning	& $41.6$ 	& $48.3$ 	& $7.4$ 	& $2.7$ 	& $0.0$ 	& $3.3 \pm 0.35$ \\	
Understanding task achievability	& $50.3$ 	& $38.9$ 	& $10.1$ 	& $0.7$ 	& $0.0$ 	& $3.4 \pm 0.36$ 	\\
Temporal reasoning	& $36.2$ 	& $50.3$ 	& $9.4$ 	& $4.0$ 	& $0.0$ 	& $3.2 \pm 0.34$ 	\\
Gestures and intention	& $43.0$ 	& $40.3$ 	& $13.4$ 	& $3.4$ 	& $0.0$ 	& $3.2 \pm 0.35$ 	\\
Learning from feedback	& $47.7$ 	& $39.6$ 	& $10.1$ 	& $2.7$ 	& $0.0$ 	& $3.3 \pm 0.35$ \\
\bottomrule
\end{tabular}
}
\end{table}

\paragraph{Open-ended question} In the open-ended question, participants were asked, ``Please provide any further comments on what would be required for you to trust that the robot understands the commands that you give it.'' From the variety of responses, the following themes emerged:
\begin{itemize}
\item \textbf{Confirmation by dialogue (30 responses)}. The robot could confirm through dialogue that it understands the task. The means of confirmation ranged from repeating the command, asking whether its interpretation is correct, saying whether or to what extend it understands the command,  and elaborating the question with more details.
\item \textbf{Flexibility and Training (28 responses).} The robot could be trained to be more flexible. The following skills were mentioned by the particpants: distinguishing different languages, pitches, accents, and slang; responding only to one or more pre-specified users; to autonomously identify when to clean and understand the requirements for each cleaning case; to be able to detect when a task is impossible and to remove or circumvent obstacles; remembering last cleaning on that location; understanding the relation between words (e.g. antonyms and synonyms).
\item \textbf{Empirical success (21 responses).} The participants emphasised that repeatedly fulfilling the task successfully and in user-friendly manner was sufficient.
\item \textbf{Interface (17 responses).} The interface for communicating with the robot could be changed. In particular, participants mentioned the following: suggesting activities based on routines; report on the tasks achieved and those not achieved and why; visualisations of the inner workings and plans of the robot; notifications upon task completion; working with written rather than spoken commands; an app to control or check what the robot will do.
\item \textbf{Confirmation by other means (13 responses)}. The robot could confirm by any other means, including gestures, lights, sound.
\item \textbf{Safety (7 responses)}. The robot behaves safely and does not endanger humans. The robot must not have the ability to override commands and must be able to be stopped at any time desired by the human operator. A number of users also questioned the behaviour of the robot if components malfunctioned or other unexpected circumstances arose (e.g., a malicious agent gives the robot a misleading command).
\item \textbf{Societal factors (5 responses)}. The participants raised societal issues such as privacy, security, and unemployment of the cleaning staff.
\item \textbf{Appearance (3 responses).} The robot could be given a more human look and voice.
\item \textbf{External (2 responses).} The human operator has some responsibility as well, particularly in formulating clear and concise language as well as taking care to articulate speech well.
\end{itemize}
The remaining 23 out of 149 responses mentioned they had nothing to add.

\subsection{Expert interviews}
\label{sec: experts}
Following a questionnaire targeted to the general public, we also conducted semi-structured interviews with two roboticists that have worked with the Helios UVC Robot. Before the interview, the participants are given the information on the robot including the added sensors and speech capabilities mentioned in Section~\ref{sec: cleaningteam}. During the interview, the participants were asked their opinion on the following six topics:
\begin{itemize}
    \item The application scenarios in which the UVC robot would be the most useful.
    \item User preferences for dialogues with the robots.
    \item Challenges in training the robot, including generalisation and bias.
    \item How users could know that the robot is really language-grounded (Q4 and Q7).
    \item Requirements for additional sensors, actuators, or user interface.
    \item Safety and responsibility.
\end{itemize}
Below is a summary of their opinion.

\paragraph{Application scenarios}
Interviewee 1 mentioned environments which are visited by many people but allow a time window in which nobody is around in the room. As examples, the interviewee mentioned office environments, educational settings, sporting events, and the hospital. 

Interviewee 2 mentioned the UVC robot would be particularly useful in structured, organised environments, such that the cleaning task becomes relatively routine, and environments where hygiene is of the utmost importance. As examples, Interviewee 2 mentioned the hospital, an operation ward, a food factory, and an agricultural setting.

\paragraph{User preferences for dialogues}
Interviewee 1 mentioned simply to instruct the cleaning of a particular room, along the lines of ``Can you please sterilize room A or room B?''. The interviewee also mentioned the use of dialogues for safety checks, clarification of the task, questions on the purpose of the task (e.g. if the room had been cleaned before), and verifying task achievement. 

Interviewee 2 also mentioned the desirability of such high-level instructions such as cleaning a particular (part of) the room. Rather than instructing how to do the task, the interviewee mentioned verifying the task achievement using a visualisation of the robot's trajectory. Where possible, control of the environment and regulations are recommended instead of dialogue. However, the interviewee did mention that dialogues may be useful if there is a good reason for the robot ask questions on its own initiative, for example, to further elaborate and clarify earlier instructions, to overcome unexpected difficulties (e.g. obstacles) in task achievement, and potentially suggest better tasks than instructed by the human. The interviewee further mentioned the desirability of a GUI or gestures in addition to verbal instructions to ensure the correct place is being cleaned. Finally, the interviewee also stressed user effort as a key concern.

\paragraph{Training the robot}
Interviewee 1 stressed that the limited available data for the particular scenario (a particular user in a particular building and sets of tasks) might mean that the robot is not adequately prepared and that this may lead to safety concerns. To avoid this, extensive simulations were recommended. 

Interviewee 2 gave a differing response for the different aspects involved (visual pattern recognition, language, and navigation). For pattern recognition, the assessed risk was fairly low given that you could train the pattern recogniser first on a general data set and then fine-tune it on examples from the specific environment. For language, the interviewee mentioned that the robot may best be trained on a more limited vocabulary and then this could be sufficient to describe all relevant aspects of the robot's tasks. For the navigation, the robot could be given a map to help reduce the problem and perhaps no learning has to be involved.

\paragraph{Trust in language grounding}
As the main factors for trust in language grounding, Interviewee 1 emphasised confirmation that the robot will start the task, empirical success on the task, and the ability to verbally explain what it did and why. A combination of the user understanding the robot as well as the robot's mechanisms being interpretable, logical, and high-performing would be the primary factors.

Interviewee 2 particularly stressed empirical success on the task, giving analogous examples such as the Alexa virtual assistant developed by Amazon; the user will see when the robot does not correctly interprets the question and this reduces the trust in the language grounding. Empirical success can also be assessed before purchase, for example, by providing a video trailer of the product. The subjectivity of the user was also highlighted, particularly stressing the prior exposure to technology and what kinds of expectations the user had.

\paragraph{Additional capabilities}
Interviewee 1 considered adding additional UVC sensors across the environment to monitor disinfection levels. Similarly, motion recordings via camera or via dead reckoning system would help track the motion of the robot and see where it did not yet go. The robot can then obtain all these additional data (UVC sensors, motion data) via WiFi to inform its decision-making.

Interviewee 2 mentioned that adding further sensor or actuators would not be cost-effective but that a graphical user interface would be useful. For example, the robot could send pictures of the environment and the human can indicate which room or part of the room should be cleaned. Labels of the objects could also be shown in the GUI to avoid misunderstandings.

\paragraph{Safety and responsibility}
Interviewee 1 mentions the built-in safety feature of the UVC robot: once the UVC robot detects movement around itself, it shuts off the UVC lights. The interviewee mentioned potential limits of this feature: it has not been thoroughly tested and there may be some sensitivity issues (e.g. some objects or agents may not be detected).  The interviewee particularly stressed the importance of dialogue for the UVC robot due to the safety concerns; with ordinary vacuum cleaning robots (i.e. not based on harmful UV rays) it can be assumed that the task is completed without much risk of damaging the environment. Malicious intent could potentially be avoided by allowing only a single user who has user authentication to operate the robot. However, it is still possible that the instruction is misunderstood and the robot cannot determine that the misinterpretation is ethically wrong or unsafe in some way.

Interviewee 2 primarily stressed to limit the potential for safety violation by simplifying the environment and introducing legal frameworks. The environment can be simplified by removing objects and dynamic agents such as humans or pets. Moreover, to avoid misunderstandings, the user and robot can in advance agree on a common vocabulary for the objects and rooms that need to be cleaned. The company should provide terms and conditions to be followed by the user. The user should be fully informed on the dangers and how to use it, fill in a health and safety checklist, and potentially undergo some sort of training to prior to operating the device. The above solutions would also limit the capability of malicious intent as this would have legal consequences.

\section{Challenges for Trust in Language Grounding}
\label{sec: future}
Having surveyed the relevant technologies and key trust factors,  questionnaire and interview have helped to shed some light onto the factors for trust in language grounding in the case study. Participants mentioned various desirable capabilities, especially emphasising the ability to detect when to clean, multiple users, dialects, and languages. Beyond the cognitive capabilities of the robot, a point that came back repeatedly in the questionnaire was to limit the autonomy of the robot in terms of the commands it takes and the restrictions in terms of safety, privacy, and societal impact. We now turn to integrating these results with the literature and the wider context of human-robot teams with the aim of outlining promising research directions.

\subsection{Safety}
Safety was one of the primary trust factors mentioned in the empirical results. While safety is particularly relevant to some applications such as bomb disposal, medical recommendation, and assisted surgery, the concern is important in general when using robots. We mainly envision two ways to provide improved safety profiles.

First, to achieve safety, restricting the autonomy of the robot was an important factor mentioned  by interviewed experts in the cleaning robot case study. Future work on safety therefore is partly up to manufacturers, lawmakers, and the staff using the robot. The manufacturer should explicitly and clearly define safe usage in the user manual and should pre-program the activation of safety features such as emergency stops or pauses after the robot detects an emergency based on an emergency signal by the user or local sensory data. The staff should follow the directions of safe use given in the manual and perform risk assessments to identify potential hazards and the response to them. Lawmakers can also aid by requiring safety features to be installed around sites where robots are operating and by specifying legal consequences for not adhering to safe practice guidelines.

Second, while learning from user experience appears to be an important dimension of trust in language grounding, the process of learning and exploring the environment needs to be safe. Such guarantees are non-trivial even in the work of safe reinforcement learning. While recent works have demonstrated safe exploration in terms of stability guarantees \cite{Berkenkamp2017}, regret bounds for performance and constraint cost \cite{Efroni2020}, and high-probability constraint-satisfaction \cite{Bossens2021}, a remaining challenge is to implement such systems for language grounding contexts. Avoiding dangerous side-effects of ambiguous communications and understanding task achievability will be an important part of solving this challenge. Similarly, one of the challenges in human-robot interaction is to detect when the human is combative \cite{Thomason2019}.

Third, a potential danger is when the human has an inappropriate level of trust, which leads to over- or underestimating the capabilities of the robot and therefore the human-robot team will be inefficient or even unsafe in its operation. This challenge is being addressed by trust calibration methods through explainable and uncertainty-aware AI techniques, with potential applications such as medical recommendation and military surveillance  \cite{Naiseh2021,Tomsett2020}. Integrating explainability into AI techniques can increase trust together with improved grounding, as was demonstrated on a visual question-answering model on the COCO data set \cite{Selvaraju2019}. We believe trust calibration methods within human-robot dialogue will be a key area of research, especially for task-based dialogues as exemplified by the human-robot cleaning team case study. Unfortunately, cognitive biases exist in all phases of the pipeline, including the design and the evaluation of the AI system \cite{Bertrand2022} and mitigating these presents a significant challenge. Another challenge in this context is to allow adaptive trust calibration based on continual feedback. A first study in this regard has demonstrated adaptive trust calibration of trust in automated decision making on a pothole inspection task, where cognitive cues were presented upon weather changes that may affect the reliability of the automated decisions  \cite{Okamura2020}.

\subsection{Reasoning beyond objects: causality, temporal relations, and intentions}
Beyond objects, which can be grounded directly into sensory information, other aspects of language grounding are significantly more challenging \cite{Matuszek2018a}. Empirical findings in this paper support the importance of causal reasoning, temporal reasoning, and intentions for trust in language grounding. Within these areas, we propose two research directions.

First, due to importance of causal reasoning, causal inference may provide an interesting AI technology for further research in trust in language grounding. If large language models can already score well on tests for causal reasoning over texts, then do we really need further advancements? We do believe so. As mentioned by Feder et al. (2021) \cite{Feder2022}, who provide a detailed call for the integration between causality estimation and NLP, argue that this is needed for at least two reasons: first, there is a need to analyse texts to distinguish causal relations from spurious correlations; second,  robustness to irrelevant features (e.g. the use of different markers for filling in checklists) is one of the key concerns for improved trustworthiness. Therefore, a few exciting research directions are to integrate causality estimation methods with NLP, for example based on Bayesian networks of causality \cite{Cooper1999}, Pearle's do-calculus \cite{Pearl2018}, or Shoelkopf's Structural Causal Modelling \cite{Scholkopf2019}. In Bayesian networks, nodes represent random variables and directed edges represent conditional probabilities within an directed acyclical graph. In Pearle's \textit{do}-calculus , one assesses the probability conditioned on an \textit{intervention} setting the variable to a particular value. In Structural Causal Models, one expresses conditional probabilities as deterministic functions taking as input the parents in the causal graph and an independent noise variable , allowing to formalise an intervention as modifying either one of the random variables or one of the deterministic functions in the causal graph. Recent works have started the integration of causal estimation with neural networks \cite{Locatello2019,Goyal2021}, and therefore the application to NLP and grounded language learning seems a promising next step. Finally, an important part of the puzzle will be to automatically find the variables based on which the causal models can be formulated; this is the main topic of causal representation learning \cite{Scholkopf2019}.

Second, one may continue the development of natural language processing techniques specifically designed for reasoning in language grounding contexts. Recent works have investigated compositional reasoning \cite{Irsoy2014,Wu2021,Zhang2021}, the ability to compose complex concepts from more elementary concepts, a property considered to be essential for language grounding.  Exciting research directions going significantly beyond grounding objects include the causality of action verbs \cite{Gao2016}, temporal grounding (e.g. understanding the difference between have, had, will) \cite{Paul2017,Karch2021}, and intentional language (e.g. language with a particular goal) \cite{Mi2020,Fleischman2005}.

\subsection{The data problem}
Deep neural networks in the context of NLP have extreme numbers of parameters (in the billions), thereby requiring massive data sets for learning. Similarly, the coupling between sensorimotor grounding and language understanding often is environment-specific. Moreover, as was mentioned in the empirical study, privacy is of major concern to participants, and it provides a basic foundation of trust; at the same time, privacy, as well as security and accessibility, often provide strong constraints on the data that can be gathered. Consequently, these above factors imply that the data is often insufficient for grounding; for example, the map of the environment or the data set coupling the language corpus with available sensorimotor data may be limited or differ significantly from the target environment. Future work on addressing the data problem may focus on three key directions. 

First, rather than presenting data directly in the manner as they are collected, active learning could be leveraged to select the order of data for improved learning. One form of active learning in this sense is to select clarifying queries based on those topics that maximise the information gain and/or are particularly relevant. Thomason et al. (2017) \cite{Thomason2017} present an approach for opportunistic active learning, in which a robot is given a task through dialogue and can then opportunistically selects queries about the task by asking about objects perceived within its sensory range. The approach was shown on an object identification task to yield better long-term performance as well as providing a more fun and engaging interaction with the human due to the use of off-topic questions. In a follow-up study, the opportunistic active learning approach was combined with sensory data for language grounding \cite{Thomason2019}. A related form of active learning which may also be explored is to actively design a curriculum that will best select the new task to be commanded to the robot based on the tasks it is already able to solve. This challenge is very much within the interactive learning experiments shown in the BabyAI environment \cite{Chevalier-Boisvert2019}, where it was shown how the sample efficiency could be greatly improved by starting with a base data set and gradually incrementing the data set with demonstrations of tasks that are at the time considered too difficult to solve for the learner.

Second, one may transfer the model learned from one situation to another, across different robot platforms, possibly with different modalities, different datasets, or different applications. Recent RL works for language grounding have gone in this direction, though not yet fully realising the ideal. For example, the model-based RL approach of Narashimhan (2018) \cite{Narasimhan2018} couples the meaning of text to the transition dynamics and rewards, such that an autonomous agent can effectively bootstrap policy learning on a new domain given its description. Another example RL system for language grounding by Hermann et al. (2017) demonstrated zero-shot transfer to novel instructions, improved the speed of language acquisition grows as the learned semantic knowledge increases, and with the help of a curriculum it could learn different tasks \cite{Hermann2017}. Similarly, a variety of promising neural network approaches have been exploring few-shot learning \cite{Johnson2017,Gu2020,Brown2020,Lake2018}. The remaining work to be done is to investigate these approaches within rich language grounding contexts and to ensure their benefit for trust in human-robot teams.

Third, as was clear in the robot cleaning example, the availability of detailed maps of the environment is often assumed by roboticists while this is not always the case. In general, the environment will be partially observable and a priori unknown, since not all data about local sites will be made publicly available due to privacy and security concerns. A recent approach by Walter et al. \cite{Walter2022} exploits environment information contained in natural language utterances to update the belief distribution over possible semantic maps; this belief distribution can then be used for effective planning for tasks such as navigation and manipulation.

\subsection{Maximally interactive dialogue systems}
In human-robot teams, dialogue systems need to be maximally interactive with the environment and the user to support effective, grounded communication. 
Mixed initiative dialogue systems, such as SUGILITE \cite{Li2020a}, and methods that ask for elaboration of the instruction provided by the user \cite{Zhang2021b} fulfill some of the concerns raised in the empirical study, e.g. on the importance of clarifying ambiguity and understanding and discussing the task at hand. Learning from user experience was also mentioned as being important for trust in language grounding, and rather than relying on prior knowledge about other users obtained from data sets or Mechanical Turk, learning in online fashion from the intended user would allow for much more personalisation. To this end, the robot may learn directly from dialogue with the intended user. For example, Riou et al. (2021) proposed a framework for joint online learning of the semantic parser and the dialogue manager, based on bandits and/or reinforcement learning that are trained by interactions with an expert where the task is to identify the purpose of a picture. Further, in addition to supporting multimodal dialogue systems and techniques for sentiment analysis and social signal processing, a further challenge is to take this one step further and start incorporating bio-feedback based on markers related to trust (e.g. EEG and GSR \cite{Wang2019,Akash2018,Hu2016}), and to allow the robot to adapt maximally to these, for instance with reinforcement learning. Finally, there are a variety of recommendations for fully social robots, including also the ability to identify new users, to remember past interactions, and to personalise the interaction to the user \cite{Leite2013}. As in the work of Chow et al. (2022) \cite{Chow2022}, full conversation histories may be compressed into semantic representations, and based on this representation as a state, the dialogue manager can select the most engaging dialogues by varying diverse possible utterances based on the context. Making such dialogue managers more sensitive with regard to personality, intent, and mood would provide a window of opportunity for trustworthiness and engagement. As a cautionary note on the above, when incorporating bio-feedback and personalisation, care should be taken not be too invasive, as this is unethical, but also to clearly present the procedure and its risks to privacy to provide an appropriate level of trust.

\subsection{Measuring trust in language grounding: beyond self-reports}
As an initial exploratory work on this topic, the empirical part of the present study consisted of a questionnaire and interviews to highlight important factors. These methodologies have their limitations as they assume the self-reported scores correspond to the true score of trust in language grounding.
Alternatives for future study include gaze \cite{Hergeth2016}, facial expressions, heart rate, and voice tracking \cite{Khalid2016}, physiological measures such as EEG and GSR \cite{Wang2019,Akash2018,Hu2016}, which have been demonstrated to be useful for assessing trust within HRI or autonomous systems contexts. Other evaluations may also include hormone levels associated with trust such as elevated levels of oxytocin and lowered levels of testosterone \cite{Berends2021,Bos2010,Zak2005}, although these would be much more difficult to probe quickly. One may also evaluate trust in language grounding without special hardware but relying purely on observable behaviours, for example, how often the human relies on the robot's utterances, what level of automation is taken, etc. The caveat to heed in the above is that these measures are not specific to language grounding, so a careful coupling of scenarios relevant for language grounding must be presented.

\section{Conclusion}
\label{sec: conclusion}
Human-robot teams communicating with natural language are challenged by the problem of language grounding. Recent advances in natural language processing have made significant progress in grounding language in terms of real-world referents, and this facilitates the development of human-robot teams in which robots can be instructed to do difficult tasks without human supervision. Research into language grounding is supported by AI techniques, data sets, simulation platforms, and user interfaces, each of which are to some extent dependent on the robot platform. We hypothesise 6 language capabilities as relevant factors for trust in language grounding and confirm these trust factors on a human-robot cleaning team. To conclude the study, we propose future research in four domains: 1) ensuring the safety of human-robot teams by careful robot design, legislation, and manufacturing on the one hand, and techniques for safe AI on the other hand; 2) more advanced language grounding towards deeper understandings of causality, temporal relations, and abstract concepts; 3) further work to improve the ability of AI systems to cope with limited and out-of-distribution data; 4) the development of mixed-initiative multimodal dialogue systems that are developed from user experience and sensitively adapt to human emotions without being invasive; and 5) evaluation methods for trust in language grounding beyond those based on user reports.

\section*{Acknowledgements}
This work has been supported by the UKRI Trustworthy Autonomous Systems Hub, EP/V00784X/1, and was part of the Trustworthy Human-Robot Teams project. We are especially grateful to Dominic Price who made the videos for the questionnaires and to Paurav Shukla for the Prolific setup.

\bibliographystyle{IEEEtran}
\bibliography{library}

\end{document}